\documentclass{article}

\usepackage{microtype}
\usepackage{graphicx}
\usepackage{subcaption}
\usepackage{caption}
\usepackage{booktabs} 
\usepackage{multirow}

\usepackage{siunitx}
\usepackage{hyperref}
\usepackage{enumitem}

\usepackage[accepted]{icml2026}

\usepackage{amsmath}
\usepackage{amssymb}
\usepackage{mathtools}
\usepackage{amsthm}

\usepackage[capitalize,noabbrev]{cleveref}

\theoremstyle{plain}

\theoremstyle{definition}

\theoremstyle{remark}

\usepackage[textsize=tiny]{todonotes}

\icmltitlerunning{Uncertainty-Calibrated Spatiotemporal Field Diffusion with Sparse Supervision
}

\begin{document}

\twocolumn[
  \icmltitle{Uncertainty-Calibrated Spatiotemporal Field Diffusion with Sparse Supervision}



  \icmlsetsymbol{equal}{*}

  \begin{icmlauthorlist}
\icmlauthor{Kevin Valencia}{xxx}
\icmlauthor{Xihaier Luo}{yyy}
\icmlauthor{Shinjae Yoo}{yyy}
\icmlauthor{David Keetae Park}{yyy}
\end{icmlauthorlist}

\icmlaffiliation{xxx}{University of California, Los Angeles}
\icmlaffiliation{yyy}{Brookhaven National Laboratory}

\icmlcorrespondingauthor{David Keetae Park}{dpark1@bnl.gov}

  \icmlkeywords{Machine Learning, ICML}

  \vskip 0.3in
]



\printAffiliationsAndNotice{}  

\begin{abstract}
Physical fields are typically observed only at sparse, time-varying sensor locations, making forecasting and reconstruction ill-posed and uncertainty-critical. We present SOLID, a mask-conditioned diffusion framework that learns spatiotemporal dynamics from sparse observations alone: training and evaluation use only observed target locations, requiring no dense fields and no pre-imputation. Unlike prior work that trains on dense reanalysis or simulations and only tests under sparsity, SOLID is trained end-to-end with sparse supervision only. SOLID conditions each denoising step on the measured values and their locations, and introduces a dual-masking objective that (i) emphasizes learning in unobserved void regions while (ii) upweights overlap pixels where inputs and targets provide the most reliable anchors. This strict sparse-conditioning pathway enables posterior sampling of full fields consistent with the measurements, achieving up to an order-of-magnitude improvement in probabilistic error and yielding calibrated uncertainty maps ($\rho > 0.7$) under severe sparsity.
\end{abstract}

\section{Introduction}

\begin{figure}[t]
\begin{center}
\includegraphics[width=1.0\columnwidth]{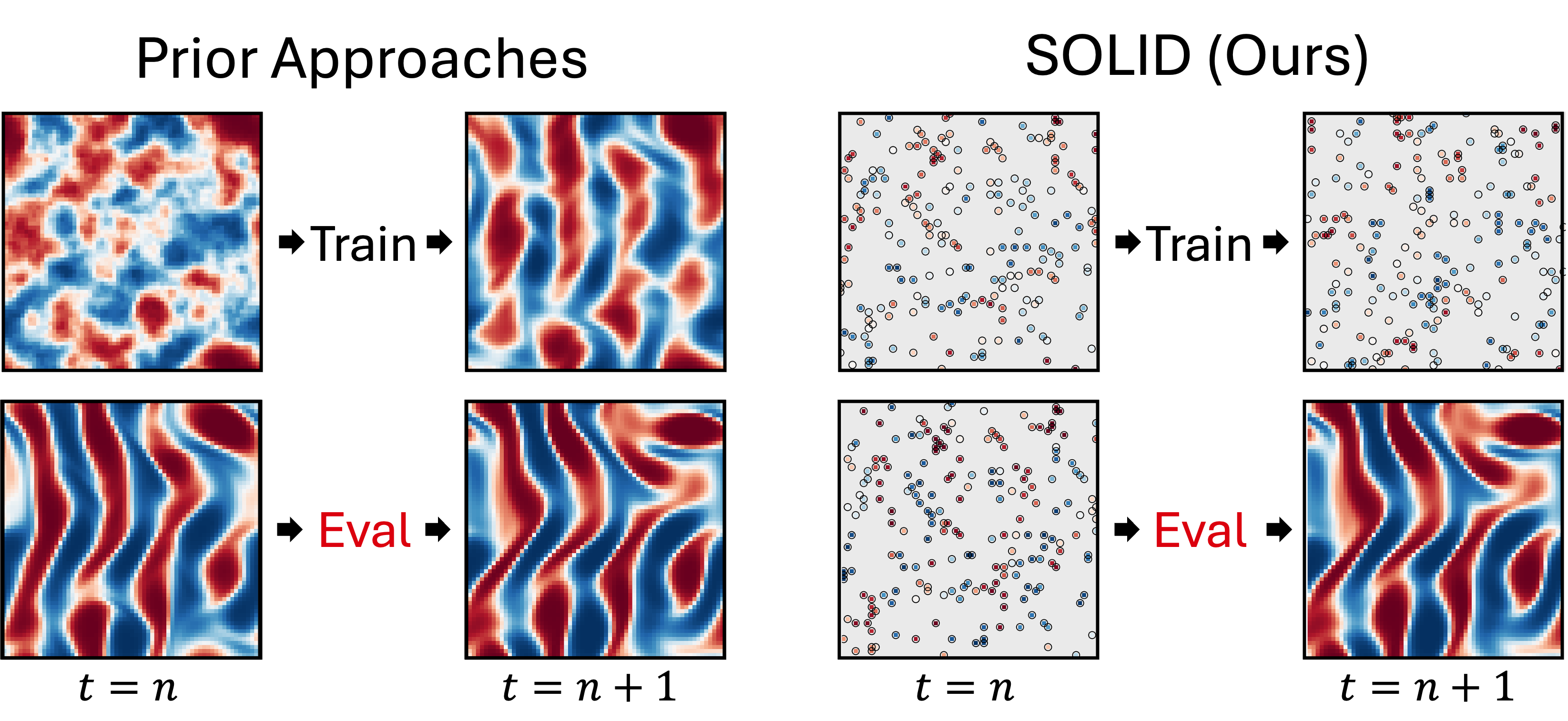}
\caption{\textbf{Conditional Field Forecasting.} This paper develops realistic training scheme with sparse data \emph{only} which contrasts with prior works assuming fully observed spatiotemporal fields. We evaluate capacities on spatiotemporal field forecasting during the inference learned \emph{purely} from sparse data. 
}
\label{fig:fig1_schematic}
\end{center}
\vspace*{-0.25in}
\end{figure}

Modeling how fields evolve over space and time is central to problems spanning climate and atmospheric dynamics, fluid transport, seismic inversion, and dynamic medical imaging~\cite{cressie2011statistics,rasmussen2006gaussian,kalnay2003atmospheric,evensen2009data,lustig2007sparse}. Progress here promises tangible impact: better forecasts, safer operations, and accelerated discovery. However, a fundamental gap exists between the continuous nature of physical phenomena and the discrete way we observe them. In practice, observations are often \emph{sparse}: only a small subset of space--time locations are measured, leaving most of the field unobserved~\cite{rasmussen2006gaussian}. This sparsity arises from physical constraints, limited sensing coverage, bandwidth and storage limits, or economic cost, and it fundamentally changes the learning problem. Learning dynamics from sparse measurements is a severely \emph{ill-posed inverse problem}: many high-dimensional fields can agree with the same limited observations~\cite{tikhonov1977solutions, tarantola2005inverse, stuart2010inverse}. Consequently, a single point estimate can be brittle and misleading, especially in scientific and mission-critical settings. These applications demand not only predictions, but also \emph{calibrated uncertainty quantification} to assess risk, build trust, and support downstream tasks such as data assimilation and adaptive sensing~\cite{gneiting2007strictly, kendall2017uncertainties, gal2016dropout}.

Current methodologies typically tackle this challenge from two disconnected perspectives: data-centric preprocessing and model-centric architectures.
\begin{itemize}[itemsep=0.5\baselineskip, topsep=0.5\baselineskip]
    \item \textbf{\textit{(P1) Data-Centric Strategies.}} A common approach is to \emph{densify} sparse observations before modeling, e.g., by interpolation or (ensemble-)variational data assimilation that produces gridded analyses from incomplete measurements~\cite{gandin1963objective, matheron1963principles, kalnay1996ncep, hersbach2020era5, bannister2017envar}. While operationally important, these pipelines can create a surrogate target used for downstream learning: they smooth fine-scale structure, propagate sampling biases, and frequently fail to propagate observational uncertainty through the full forecasting stack when deterministic analyses are treated as ground truth~\cite{stein1999interpolation}. More recent approaches---including set-to-function models such as conditional neural processes and Fourier neural processes, as well as masked pretraining---can learn flexible priors over missing values~\cite{garnelo2018neural, kim2019anp, chen2024fnp, reed2023scale}, but are frequently used as a separate pretext stage: an imputed dense field is still fed to a downstream predictor, reintroducing the information bottleneck and discarding uncertainty where it matters most.
    
    
    \item \textbf{\textit{(P2) Model-Centric Strategies.}} Many forecasting architectures implicitly assume dense inputs. Grid-based CNNs and Transformers typically require complete tensors, making performance sensitive to missingness and encouraging ad-hoc imputation or masking heuristics~\cite{krizhevsky2012imagenet, dosovitskiy2021vit, pathak2022fourcastnet}. Neural field approaches provide a promising alternative by representing fields continuously and enabling queries at arbitrary coordinates~\cite{sitzmann2020siren, li2020fno, lu2021deeponet}; moreover, NeRF/INR-inspired models can be adapted to sparse regimes via additional structure and optimization (e.g., dynamical reconstruction from sparse observations)~\cite{saraertoosi2025neuraldmd}. However, these approaches are often used deterministically or without calibrated posterior sampling, leaving uncertainty under-modeled in the sparse-observation regime that motivates our method.
\end{itemize}

The preceding analysis reveals a critical gap: \textit{data-side pipelines often hallucinate missing structure while collapsing uncertainty, and model-side approaches frequently assume dense observations or provide limited probabilistic inference under sparsity}. This motivates a unified probabilistic approach that (1) learns directly from sparse observations without requiring dense pre-imputation, (2) frames forecasting as inverse inference under missingness, and (3) delivers calibrated predictive uncertainty.
Overall, our contributions are three-fold:

\begin{itemize}[itemsep=0.25\baselineskip, topsep=0.25\baselineskip]
\item \textbf{SOLID.} A sparsity-aware diffusion that generalizes across regimes and patterns.
SOLID (\emph{\underline{S}parse-\underline{O}n\underline{L}y f\underline{I}eld \underline{D}iffusion}) couples a dual-masked denoising objective with mask-conditioned sampling, enabling a single model to operate reliably from extreme sparsity to denser settings and across diverse sparsity geometries, yielding practical guidance for sensor-network design.
\item \textbf{Performance.} SOLID consistently outperforms \emph{nine} strong baselines across a broad suite of tasks, spanning complex physics simulations and real-world benchmarks, with a demonstrated data- and parameter-efficient training.
\item \textbf{Calibrated Uncertainty.} SOLID delivers trustworthy probabilistic forecasts by generating uncertainty maps that are highly correlated with prediction error (e.g., $\rho > 0.7$), effectively learning to identify the regions that are hardest to predict.
\end{itemize}

\section{Related Work}
\label{sec:related_work}
Our work sits at the intersection of spatiotemporal modeling, representation learning, and probabilistic deep learning. We position our efforts by reviewing two paradigms for handling sparse scientific data: data-centric preprocessing and model-centric architectural choices.

\begin{figure*}[t]
\begin{center}
\includegraphics[width=0.85\textwidth]{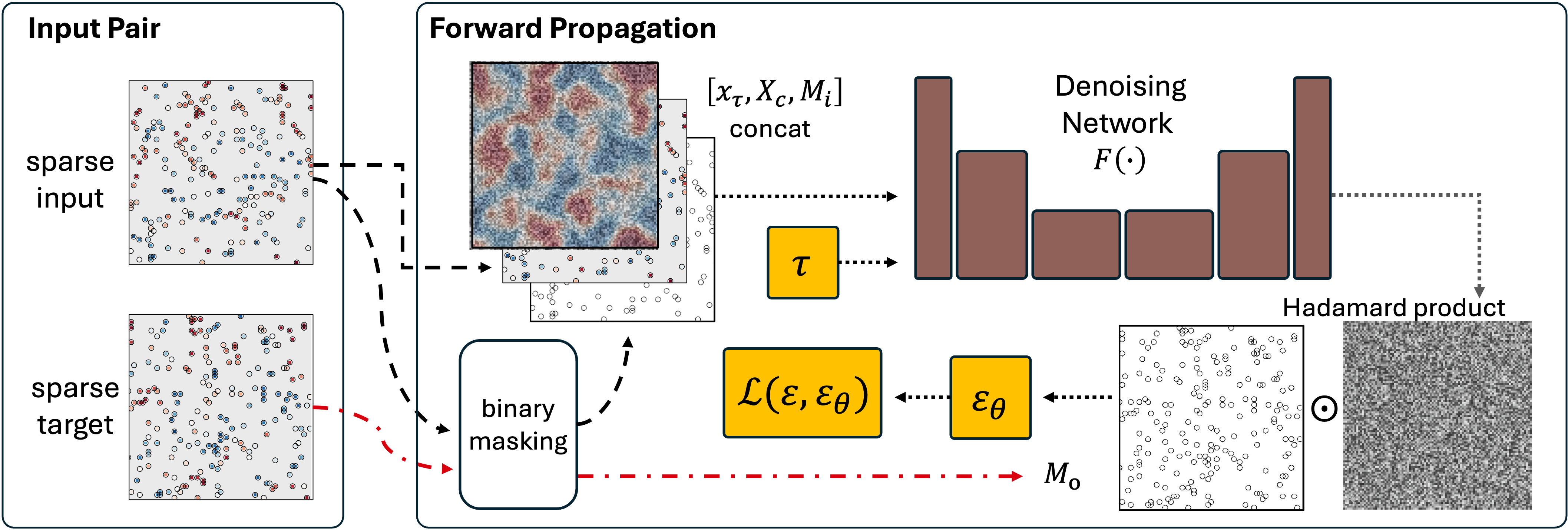}
\caption{\textbf{Training the Proposed SOLID.} A dual-masking strategy is introduced. Coupled with sparse conditioning, it realizes a simple yet effective spatiotemporal learning capable of field reconstruction robust with sparse-only training data.}
\label{fig:fig2_solid_model}
\end{center}
\vspace*{-0.2in}
\end{figure*}

\subsection{Data-Centric Strategies}

\paragraph{Classical Interpolation \& Data Assimilation.} 
The foundational approach in science is to first regularize sparse data. Methods like kriging use statistical priors, while data assimilation uses physical simulators, to project sparse observations onto a complete, regular grid~\cite{matheron1963principles, kalnay2003atmospheric,barthelemy2022super,qu2024deep,chen2024fnp}. This process creates the canonical reanalysis datasets (e.g., ERA5) that have fueled modern deep learning for weather forecasting~\cite{pathak2022fourcastnet, hersbach2020era5}. However, this creates a \textit{surrogate reality}: a single, deterministic field that is inherently over-smoothed, losing critical high-frequency details~\cite{stein1999interpolation,shysheya2024conditional,li2025frequency}, and collapses the true observational uncertainty into an artifact treated as ground truth.
\paragraph{Modern Learned Imputation.} More sophisticated data-driven interpolators have emerged, including Neural Processes (NPs) that learn distributions over functions from arbitrary points~\cite{garnelo2018neural,garnelo2018conditional}, and masked auto-encoders that learn physical priors via self-supervision~\cite{ye2022contrastive,lin2022conditional,reed2023scale}. However, their power is undermined by their typical use as a \textit{disjoint pretext task}~\cite{patel2022neur2sp,zhou2022neural}. They generate a single, dense field that is treated as deterministic truth by a separate downstream model. This two-stage pipeline creates a critical information bottleneck, severing the flow of uncertainty and allowing imputation errors to silently propagate and compound.

\subsection{Model-Centric Strategies}

\paragraph{CNNs \& Transformers.} While CNNs and Transformers have excelled on scientific datasets~\cite{krizhevsky2012imagenet, pathak2022fourcastnet}, they typically assume a lattice-structured input, so applying them to observational data often requires a pre-processing step that maps samples onto a grid via imputation. While gridding itself can be a reasonable choice, this pre-imputation can introduce smoothing and method-dependent artifacts, shifting learning away from the raw measurements~\cite{dosovitskiy2021vit,zhou2021informer,choromanskirethinking}.

\paragraph{Graph Neural Networks (GNNs).} Although GNNs accommodate sparse data, their architecture is brittle, relying entirely on a predefined and static graph topology~\cite{lam2023graphcast,wang2020traffic,wang2021end,li2021spatial}. This makes the vital task of querying arbitrary, off-graph locations fundamentally difficult. Moreover, their reliance on local message-passing is ill-suited for physical systems dominated by long-range interactions, requiring deep graphs that are both computationally demanding and prone to over-smoothing features~\cite{thurlemann2023anisotropic,wang2024neural,wu2021representing}.
\paragraph{Neural Fields.} This family of models learns continuous, functional representations of data. Specifically, neural fields map input coordinates to a value, famously used in NeRF for view synthesis~\cite{sitzmann2020siren, mildenhall2021nerf,lu2023learning,pan2023neural} and neural operators go a step further by learning mappings between infinite-dimensional function spaces, making them powerful tools for solving families of PDEs~\cite{li2020fno, lu2021deeponet,azizzadenesheli2024neural,salvi2022neural}.
 Though promising, they are deterministic, producing a single, overconfident prediction for an ill-posed inverse problem that demands a full posterior distribution. 

\paragraph{Deep Generative Models.}
Deep generative priors---notably diffusion~\cite{ho2020denoising,song2021solving} and adversarial models~\cite{baselineINRGAN,baselineAliasFreeGAN}---are widely used for ill-posed reconstruction.
Under sparsity, many pipelines still densify measurements onto grids before applying the prior, which can smooth structure and collapse uncertainty~\cite{ruhling2023dyffusion,wang2023extraction,valencia2025learning}.
More principled alternatives either (i) learn an amortized conditional generator by conditioning on values and masks during training, or (ii) impose data-consistency only at inference via posterior-sampling or projection routines (e.g., PMA-Diffusion; sequential diffusion under irregular operators)~\cite{liu2025pmadiffusion,chen2025sdift}.
SOLID follows the former, but is designed for \emph{sparse-only supervision}: it injects the masked values at every denoising step and supervises only observed targets, yielding conditional ensembles without external inference machinery.

\section{Spatiotemporal Field Forecasting}
\label{sec:method}

\subsection{Problem Statement}
\label{sec:problem_statement}

We consider an unknown spatiotemporal field $U^t \in \mathbb{R}^{H\times W}$ on a fixed grid $\Omega=\{1,\dots,H\}\times\{1,\dots,W\}$ at discrete times $t\in\mathbb{Z}$. At any $t$, a fully observed field $U^t$ is typically unavailable—real deployments measure only a sparse, time-varying subset $O_t\subset\Omega$ due to limited instrument coverage and cost, sensor outages/QA\!/QC filtering, and asynchronous or moving acquisition—so we observe $\mathcal{R}_{M_t}(U^t)=M_t\odot U^t$, where the binary mask $M_t\in\{0,1\}^{H\times W}$. We write the restriction operator $\mathcal{R}_{M}(U)=M\odot U$ (Hadamard product). Each training example provides a \emph{single-step} pair
\[
\big(\mathcal{R}_{M_i}(U^{t_i}),\,M_i\big)\quad\text{and}\quad\big(\mathcal{R}_{M_o}(U^{t_o}),\,M_o\big),
\]
with $t_o\in\{t_i,\,t_i{+}1\}$. When $t_o{=}t_i$ the task is same-time \emph{reconstruction}; when $t_o{=}t_i{+}1$ it is \emph{single-step forecasting}. Supervision is available \emph{only} on $M_o$. The goal is to learn a predictor $F$ that maps sparse input and masks to a dense estimate on the grid, i.e.,
$F\!:\big(\mathcal{R}_{M_i}(U^{t_i}),M_i,t_i,t_o\big)\mapsto \hat U^{t_o}\in\mathbb{R}^{H\times W}$,
while training losses are evaluated exclusively on $M_o$; no full fields are required at train or test time.
\vspace{-0.5em}

\subsection{SOLID}
\label{sec:method_sparse}

In this section, we introduce SOLID (Fig.~\ref{fig:fig2_solid_model}). 

\paragraph{Denoising Diffusion Model.} Denoising diffusion models~\cite{ho2020denoising} learn a generative Markov chain that gradually transforms Gaussian noise into a sample from the data distribution. Let \(x_0\) denote the clean target we wish to generate; the \emph{forward} (noising) process constructs a sequence \(\{x_\tau\}_{\tau=1}^T\) by adding Gaussian noise with a prescribed variance schedule \(\{\beta_\tau\}\), yielding the closed form:
\begin{equation}
\label{eq:diffusion_model}
x_\tau \;=\; \sqrt{\bar\alpha_\tau}\, x_0 \;+\; \sqrt{1-\bar\alpha_\tau}\, \varepsilon,\qquad \varepsilon\sim\mathcal N(0,I),
\end{equation}
where \(\alpha_\tau=1-\beta_\tau\) and \(\bar\alpha_\tau=\prod_{s=1}^{\tau}\alpha_s\). In our paper, \(x_0\) is the spatiotemporal field at the supervision time \(t_o\) (i.e., \(x_0=U_{t_o}\)) and we adopt exactly this standard parameterization. The \emph{reverse} process is learned by a denoiser \(\varepsilon_\theta(x_\tau,\tau,\cdot)\) that predicts the noise injected at each step. The common \textit{\(\varepsilon\)-prediction} objective minimizes mean-squared error between the true \(\varepsilon\) and the network’s prediction. In SOLID we use the same objective but restrict supervision to locations where targets exist (see masked objective below), matching Eq.~\ref{eq:diffusion_model}.

\paragraph{Conditioning.}
From the input slice we form $X_c = \mathcal{R}_{M_i}(U^{t_i}) = M_i \odot U^{t_i}$. At diffusion step $\tau\in\{1,\dots,T\}$ the denoiser receives the channel-wise concatenation $[x_\tau,\,X_c,\,M_i]\in\mathbb{R}^{3\times H\times W}$, where $x_\tau$ is the noised target $x_0{=}U^{t_o}$ using the usual forward process $x_\tau=\sqrt{\bar\alpha_\tau}\,x_0+\sqrt{1-\bar\alpha_\tau}\,\varepsilon$ with $\varepsilon\sim\mathcal{N}(0,I)$. The time index $\tau$ is embedded by a sinusoidal encoding followed by a multi-layer perceptron (MLP) block.

\paragraph{Dual-Masked Denoising Objective.}
We supervise \emph{only} where targets exist. Let $M_o\in\{0,1\}^{H\times W}$ denote the target mask at $t_o$, $X_c=M_i\odot U^{t_i}$ the sparse input, and $[x_\tau,X_c,M_i]$ the denoiser input at diffusion step $\tau$. With $\varepsilon\!\sim\!\mathcal{N}(0,I)$ and $\varepsilon_\theta$ the UNet~\cite{baselineUNet}, the loss is
\begin{equation}
\label{eq:masked_loss_simple}
\mathcal{L}(\theta)\;=\;\mathbb{E}_{\tau,\varepsilon}\!\left[
\frac{\big\|\,M_o\odot\big(\varepsilon-\varepsilon_\theta([x_\tau,X_c,M_i],\tau)\big)\big\|_2^2}{\|M_o\|_1}
\right],
\end{equation}
which never touches unobserved entries of $U^{t_o}$.

\paragraph{Overlap Reweighting.}
Locations that are both sensed at \(t_i\) and supervised at \(t_o\) carry the most reliable information and exhibit the lowest epistemic uncertainty \citep{rasmussen2006gaussian}. Empirically, uncertainty rises with distance from sensors (cf.\ Fig.~\ref{fig:uq_plots}). We therefore up-weight only these \emph{anchor} pixels in the denoising loss, using
\[
\widetilde M \;=\; M_o \odot (1+\lambda M_i) \;=\; M_o + \lambda\,(M_i \odot M_o),
\]
with small \(\lambda\). Because the loss is multiplied by \(\widetilde M\), the gradients are scaled by \(1+\lambda\) exactly at overlap pixels (\(M_i \odot M_o = 1\)) and left unchanged elsewhere. This makes the model learn most strongly where we have reliable measurements, which in turn pulls nearby predictions into alignment and smooths transitions across mask boundaries (fewer seams). And since only the overlap is up-weighted—not the entire input mask—the model is not encouraged to copy the input frame everywhere, so it can still learn genuine temporal changes instead of defaulting to persistence. As \(\lambda \to 0\) the conditioning signal is underused; as \(\lambda\) grows too large the dynamics are over-regularized. We find \(\lambda \in [0.05, 0.1]\) balances stability and flexibility. If there are no overlapping sensors, the \(\lambda\) term is inactive.

\paragraph{Uncertainty Map.}
We quantify predictive uncertainty by Monte Carlo sampling from SOLID under fixed conditioning. Given an input slice $(X_c, M_i)$ at time $t_i$ and a target time $t_o$, we draw $K=100$ independent conditional samples $\{\hat U_{t_o}^{(k)}\}_{k=1}^{K}$ using the same $(X_c,M_i)$ and compute the per-pixel standard deviation
\begin{align}
\label{eq:uq}
\sigma_{p}(t_o)
&= \sqrt{\frac{1}{K-1}\sum_{k=1}^{K}\Big(\hat U_{t_o}^{(k)}(p)-\bar U_{t_o}(p)\Big)^2}
\end{align}
which we report as the \emph{uncertainty map} (denoted as STD in Fig.~\ref{fig:uq_plots} and Fig.~\ref{fig:uq_qualitative}). For multi-step horizons $t_{i+h}$, $h=1,\dots,M$, we perform $K=100$ autoregressive rollouts. At each step $h$, we form the next-step input by restricting the model's previous prediction to the available input locations and reconditioning on the same mask,
\[
X_c^{(h)} \;=\; R_{M_i}\!\big(\hat U_{t_{i+h-1}}^{(k)}\big), \quad
\hat U_{t_{i+h}}^{(k)} \sim \texttt{SOLID}\big(\,\cdot\,\big|\,X_c^{(h)}, M_i\big),
\]
and treat $\{\hat U_{t_{i+h}}^{(k)}\}_{k=1}^{K}$ as the ensemble for horizon $h$. The uncertainty map at each horizon is computed by the same per-pixel standard deviation as above.







\begin{figure*}[t]
\vspace*{-0.15in}
\begin{flushright}
\begin{center}
\includegraphics[width=0.925\textwidth]{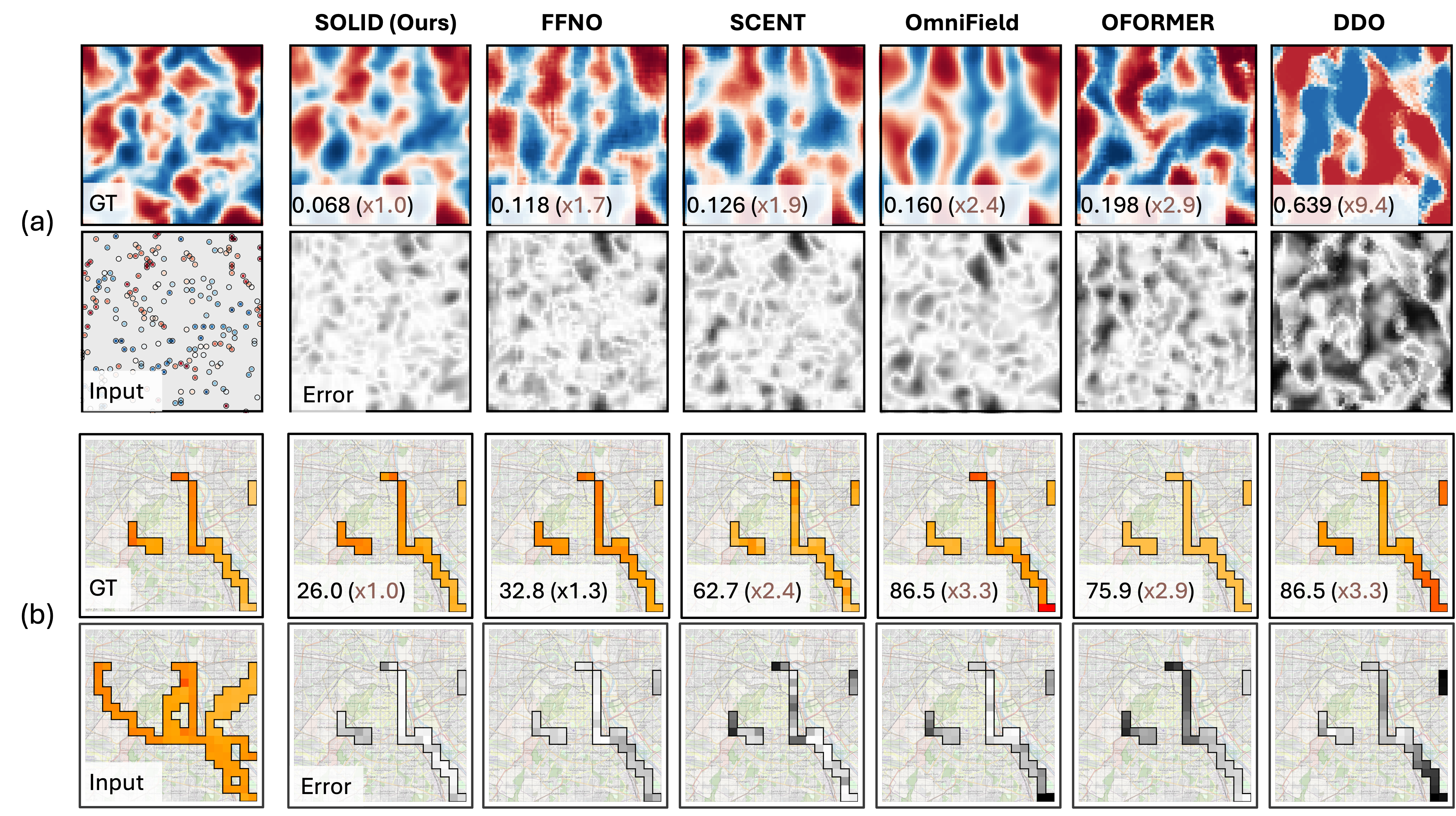}
\caption{\textbf{Qualitative Comparisons on Forecasting.} White boxes denote MSE between prediction and target GT, and gray-scale images show corresponding error.}
\label{fig:qualitative}
\end{center}
\vspace*{-0.25in}
\end{flushright}
\end{figure*}

\section{Experiments}
\label{sec:experiments}

\subsection{Experimental Setup}
\label{sec:exp_setup}
\paragraph{Datasets.}
We evaluate the proposed SOLID method against nine baseline approaches on both challenging simulated and complex real-world datasets. Details of dataset generation and preprocessing are provided in the Appendix.
\begin{itemize}[itemsep=0.25\baselineskip, topsep=0.25\baselineskip]
    \item \textbf{Navier-Stokes Dataset.} This is a dataset of 1000 2D Navier-Stokes simulation trajectories (25 timesteps each). To simulate sparse sensing, we apply sparse ma sks to this dense data, exploring both random and structured ($8 \times 8$ block) patterns. These masks are either instance-specific or shared across the dataset. Unless specified otherwise, we use fixed, instance-specific masks with 10\% randomly observed entries.
    \item \textbf{AirDelhi Dataset.} AirDelhi is a real-world benchmark of sparse PM2.5 measurements from mobile sensors on buses in Delhi, India~\cite{airdelhi}. Its sparsespatiotemporal coverage naturally creates a challenging sensor-to-sensor forecasting task. We use the standard AirDelhi Benchmark (AD-B) setup: data is aggregated into a $1km^{2}$ spatial grid and 30-minute intervals, with standard exclusions and train/test splits.
\end{itemize}

\paragraph{Baselines.}
We compare against nine baselines spanning three families: \textbullet\ \textbf{Grid-centric}: UNet~\cite{baselineUNet}. \textbullet\ \textbf{Deterministic operators}: FFNO~\cite{baselineFFNO}, OFormer~\cite{baselineOFormer}, OmniField~\cite{baselineOmniField}, and SCENT~\cite{baselineSCENT}. \textbullet\  \textbf{Generative models}: StyleGAN3~\cite{baselineAliasFreeGAN}, INR-GAN~\cite{baselineINRGAN}, $\infty$-Diff~\cite{baselineInftyDiff}, and DDO~\cite{baselineDDO}. Some prior diffusion methods handle sparse observations via inference-time posterior sampling with explicit physics constraints~\cite{liu2025pmadiffusion,chen2025sdift}. 
They assume an explicit forward operator and projector-style inference, and are thus not directly comparable to our amortized forecasting setting with sparse supervision. For fair comparisons across baselines, sparse data were given without preprocessing or masking, yet target masks $M_o$ were provided during training whenever feasible.


\paragraph{Metrics.}
Our primary evaluation metric is the Continuous Ranked Probability Score (CRPS), a proper scoring rule that evaluates the entire predictive distribution for both sharpness and calibration. CRPS evaluates the entire forecast distribution \(F\) at each location $X$ against the value \(y\) via
\begin{align}
\small
\mathrm{CRPS}(F,y)&=\int_{-\infty}^{\infty}\!\big(F(z)-\mathbf{1}\{z\ge y\}\big)^2\,dz \;\\&=\; \mathbb{E}_{X\sim F}\!\left[|X-y|\right]-\tfrac{1}{2}\,\mathbb{E}_{X,X'\sim F}\!\left[|X-X'|\right],
\end{align}
and we average this over the observed set defined by \(M_o\). Concretely, given $K$ conditional samples $\{\hat U^{(k)}_{t_o}\}_{k=1}^K$ under fixed conditioning $(X_c,M_i)$,
\begin{align*}
\small
\widehat{\mathrm{CRPS}}_{t_o}
&= \frac{1}{\|M_o\|_1}
\sum_{p\in\Omega} M_o(p)\Bigg[
\overbrace{\frac{1}{K}\sum_{k=1}^{K}\!\big|\hat U^{(k)}_{t_o}(p)-U_{t_o}(p)\big|}^{\text{MAE (observation error)}}
\\&-\overbrace{\frac{1}{K(K-1)}\!\!\sum_{1\le k<\ell\le K}\!\!\big|\hat U^{(k)}_{t_o}(p)-\hat U^{(\ell)}_{t_o}(p)\big|}^{\text{ensemble dispersion (sharpness)}}
\Bigg].
\label{eq:crps-mc}
\end{align*}
When $K{=}1$, $\widehat{\mathrm{CRPS}}$ reduces to MAE on $M_o$, meaning that CRPS reduces to the Mean Absolute Error (MAE) for deterministic forecasts, enabling a fair comparison across all models. We also report the standard Mean Squared Error (MSE), computed as \(\text{MSE}=\frac{1}{\lVert M_o\rVert_1}\sum_{(i,j)\in M_o}\!(\hat U^{ij}-U^{ij})^2\), for its familiarity and for comparing single generated samples. 

\paragraph{Training and Inference Details.}
We train SOLID on a UNet backbone (36~M parameters) using Adam optimizer with a learning rate of $2 × 10^{-5}$ and gradient clipping (max-norm 1.0). The diffusion process follows DDPM~\cite{ho2020denoising} with $\tau=1000$ steps and MSE loss, and inference uses DDIM~\cite{song2020denoising} with 50 steps. Training is performed on an NVIDIA A100 (40~GB) GPU with a batch size of 32 for 100K steps (Appendix Table~\ref{tab:solid_hparams_appendix} for full details).

\begin{figure}[t]
\begin{center}
\includegraphics[width=0.95\columnwidth]{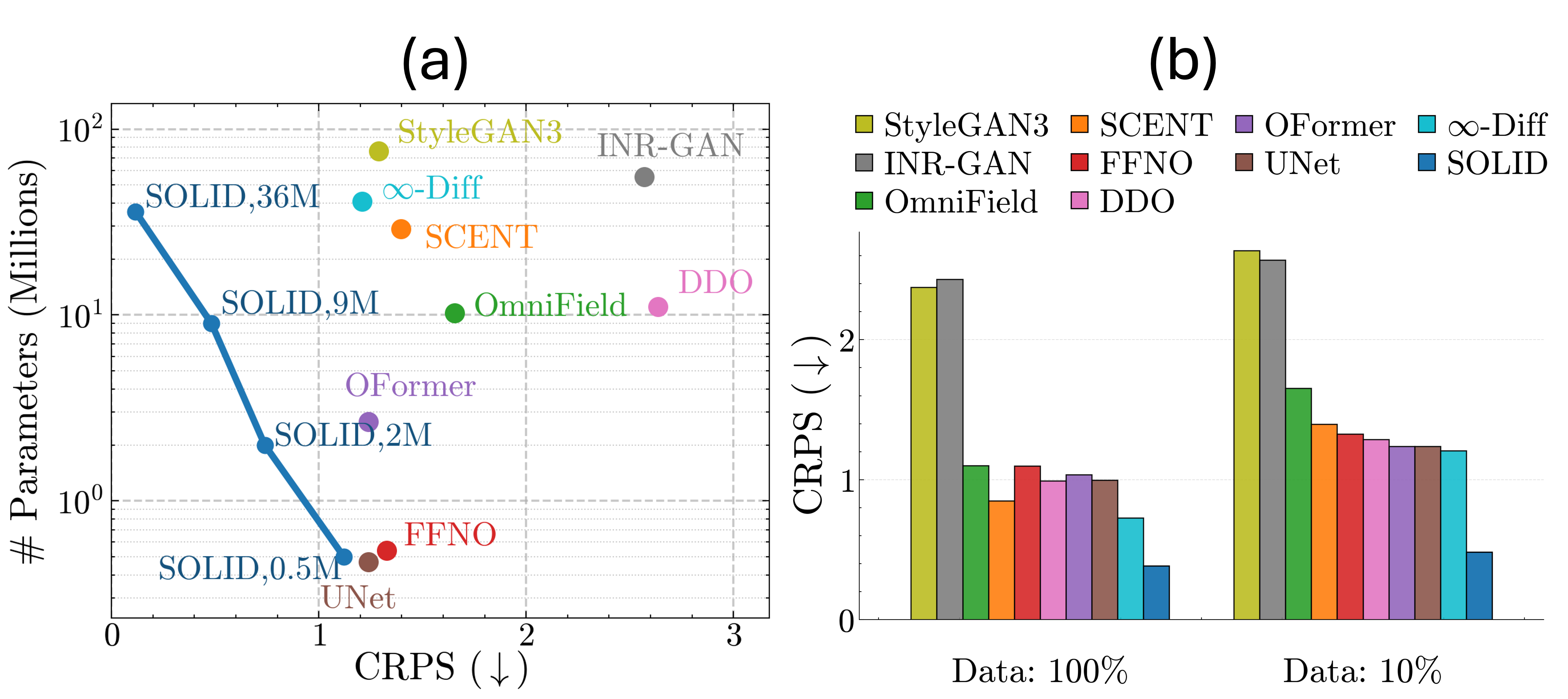}
\caption{\textbf{Performance and Parameter Efficiency.} Navier–Stokes baseline comparisons summarizing (a) Per parameter efficiency and (b) data efficiency (full data vs.10\%).}
\label{fig:efficiency}
\end{center}
\vspace*{-0.25in}
\end{figure}

\begin{figure}[t]
\begin{center}
\includegraphics[width=1.0\columnwidth]{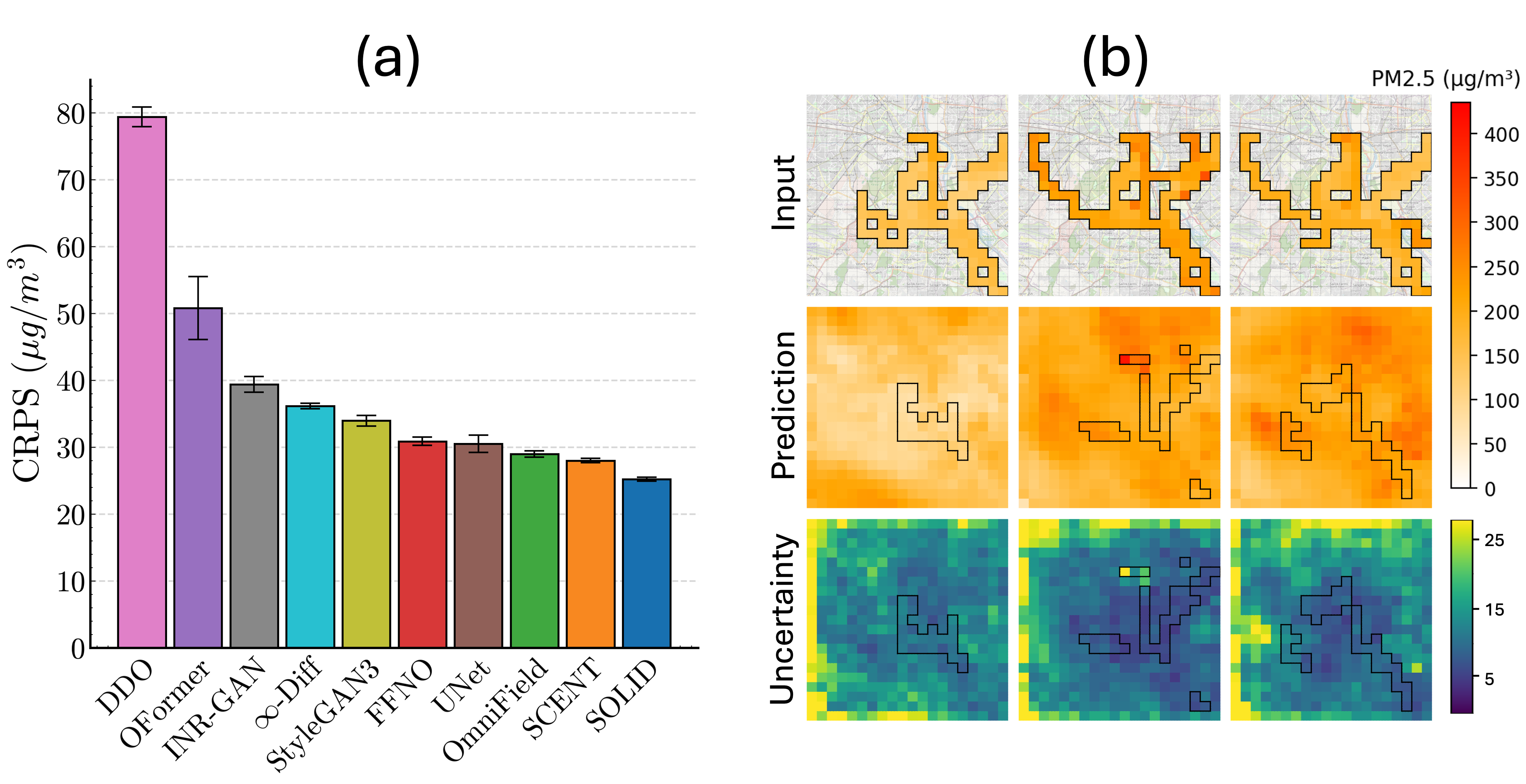}
\caption{\textbf{AirDelhi: Performance and Uncertainty.} (a) Each model is run for ten different seeds, shown as standard deviation. (b) 30-minute air pollution prediction and its uncertainty. Black contours denote available sample positions.}

\label{fig:air_delhi_perf}
\end{center}
\vspace*{-0.25in}
\end{figure}

\subsection{Task 1: Forecasting}
\label{sec:main_result}
\begin{figure}[t]
\begin{center}
\includegraphics[width=1.0\columnwidth]{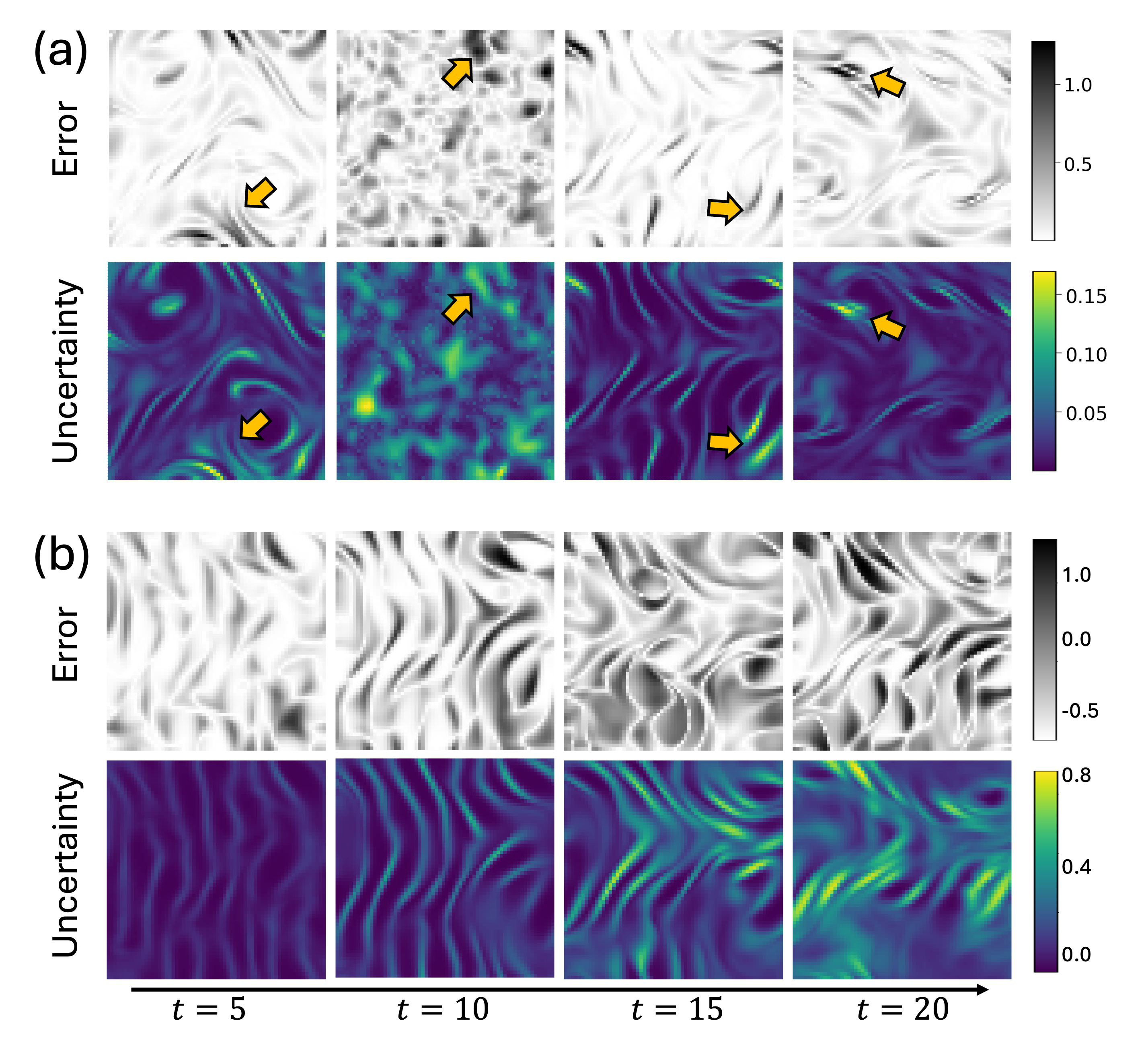}
\caption{\textbf{Qualitative Analyses of Uncertainty.} Regions of high uncertainty co-locate with (a) large error (Navier–Stokes). Orange-colored arrows point to representative areas where error and uncertainty highly match. (c) Rollouts on a single sparse input show that uncertainty propagates over time, following closely with the elevating error.}

\label{fig:uq_qualitative}
\end{center}
\vspace*{-0.3in}
\end{figure}

\begin{figure}[t]
\begin{center}
\includegraphics[width=1.02\columnwidth]{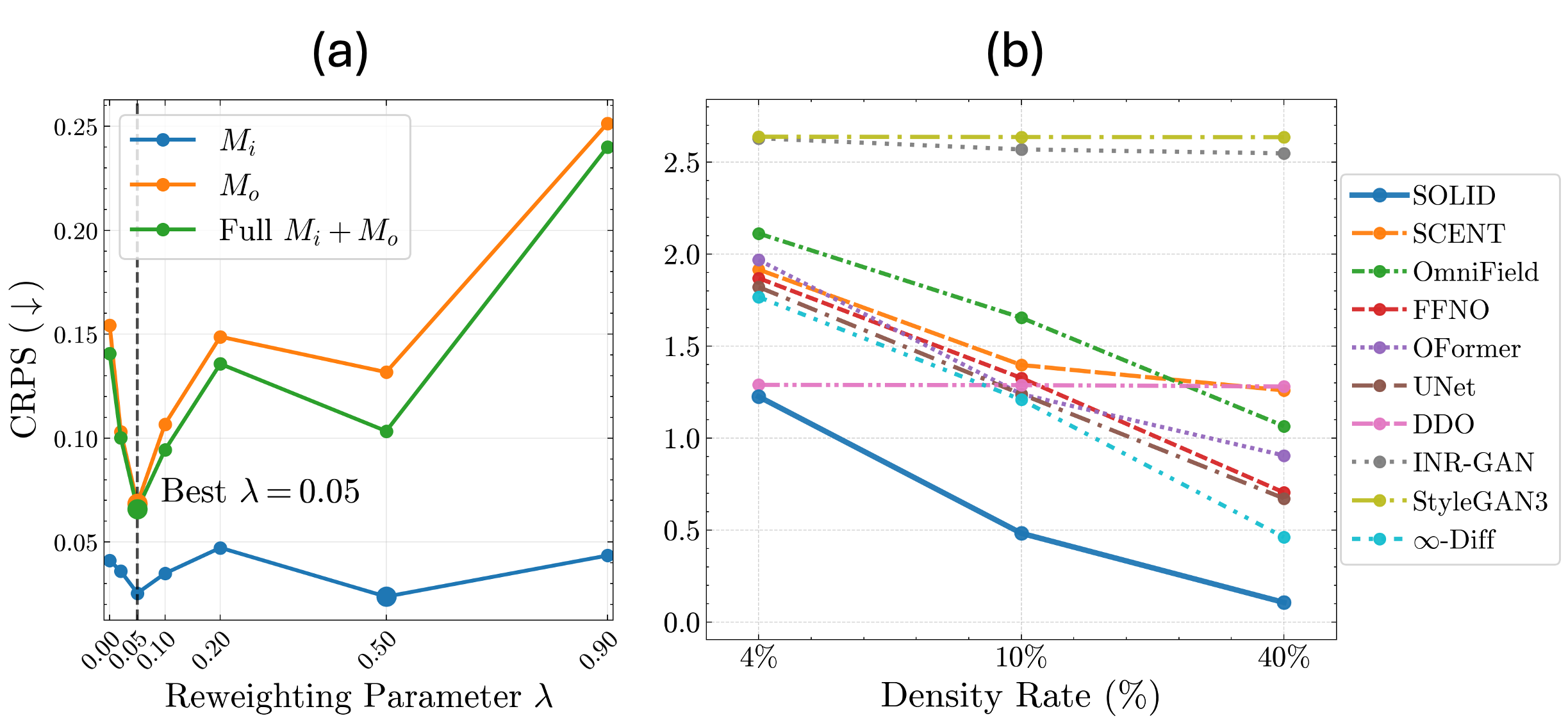}
\vspace*{-0.15in}
\caption{\textbf{Ablations on $\lambda$ and Sparsity Rate.} }
\label{fig:ablation_lambda}
\end{center}
\vspace*{-0.3in}
\end{figure}

In sensor-only inference, the core question is accuracy under uncertainty: we need predictions that are both sharp and \emph{calibrated}. We therefore report CRPS, which jointly evaluates pointwise accuracy and uncertainty calibration. On Navier–Stokes spatiotemporal field reconstruction, both qualitative (Fig.~\ref{fig:qualitative}a) and test evaluations (Fig.~\ref{fig:efficiency}a-b) show that SOLID significantly outperforms all baselines in CRPS. The gap persists across training regimes (high-data and low-data) and sparsity rates (Fig.~\ref{fig:ablation_lambda}b), indicating that SOLID’s probabilistic reconstructions remain accurate and well-calibrated where the task is most ill-posed. Moreover, longer-horizon rollouts exhibit the expected gradual error growth under sparse conditioning, and critically, SOLID’s calibrated uncertainty co-locates with errors, enabling risk-aware downstream decisions (Fig.~\ref{fig:uq_qualitative}b).

On the real-world AirDelhi benchmark, the qualitative comparisons on a sample input in Fig.~\ref{fig:qualitative}b show that SOLID reconstructs fine-scale PM2.5 structure from sparse, bus-mounted sensors with sharper fronts, while baselines either oversmooth (FFNO/OmniField), exhibit blocky artifacts (OFormer), or hallucinate spurious patterns ($\infty$-Diff/DDO); the per-panel white boxes (MSE) and the accompanying absolute-error maps consistently show lower residuals for SOLID than for competing methods, reflecting fewer large-error regions overall. These visual trends align with the quantitative probabilistic metric (CRPS). Fig.~\ref{fig:air_delhi_perf} shows that SOLID attains 25.71 CRPS vs. 28.47 for SCENT ($\downarrow$ 2.75, 9.7\% relative gain), while also consistently outperforming other baselines, demonstrating that strict mask-conditioned diffusion recovers coherent fields in the unobserved “void” where alternatives falter. 

\subsection{Task 2: Data \& Model Efficiency Study}
\label{sec:efficiency}

We sweep parameter budgets and training-trajectory counts to assess parameter- and data-efficiency (Fig.~\ref{fig:efficiency}). In the model-scaling plot (Fig.~\ref{fig:efficiency}a), SOLID traces the Pareto front from sub-million to tens-of-millions of parameters, consistently achieving lower CRPS than all competitors. Notably, although FFNO and UNet appear competitive in the smallest-model regime, scaling them up increases the gap:
FFNO (37.00M) reaches 0.704 and UNet (29.7M) reaches 1.011, versus SOLID (36M) at 0.101 in CRPS.
In the data-scaling barplots (Fig.~\ref{fig:efficiency}b), SOLID attains the lowest CRPS in both the high-data regime (1000 trajectories) and the low-data regime (100 trajectories); several baselines degrade sharply as data shrinks. Deterministic operator learners---while mesh-agnostic---remain point-estimate models and fall behind when calibrated uncertainty and strict conditioning are required.

\noindent\textbf{Compute trade-off.} While SOLID is parameter-efficient per denoising step, its probabilistic predictions incur higher inference cost (see Appendix Table~\ref{tab:flops_params_all} for FLOPs comparisons): we run 50 denoising steps per sample, and uncertainty estimation requires multiple Monte Carlo samples, making total sampling cost substantially higher than single-pass deterministic forecasters, which is a common limitation of diffusion-based predictors. Together, Fig.~\ref{fig:efficiency} supports SOLID as the most accurate and the most parameter-data-efficient approach among the compared families, with a clear accuracy--calibration versus compute trade-off at inference time.



\subsection{Task 3: Uncertainty Analysis}
\begin{figure}[t]
\begin{center}
\includegraphics[width=0.975\columnwidth]{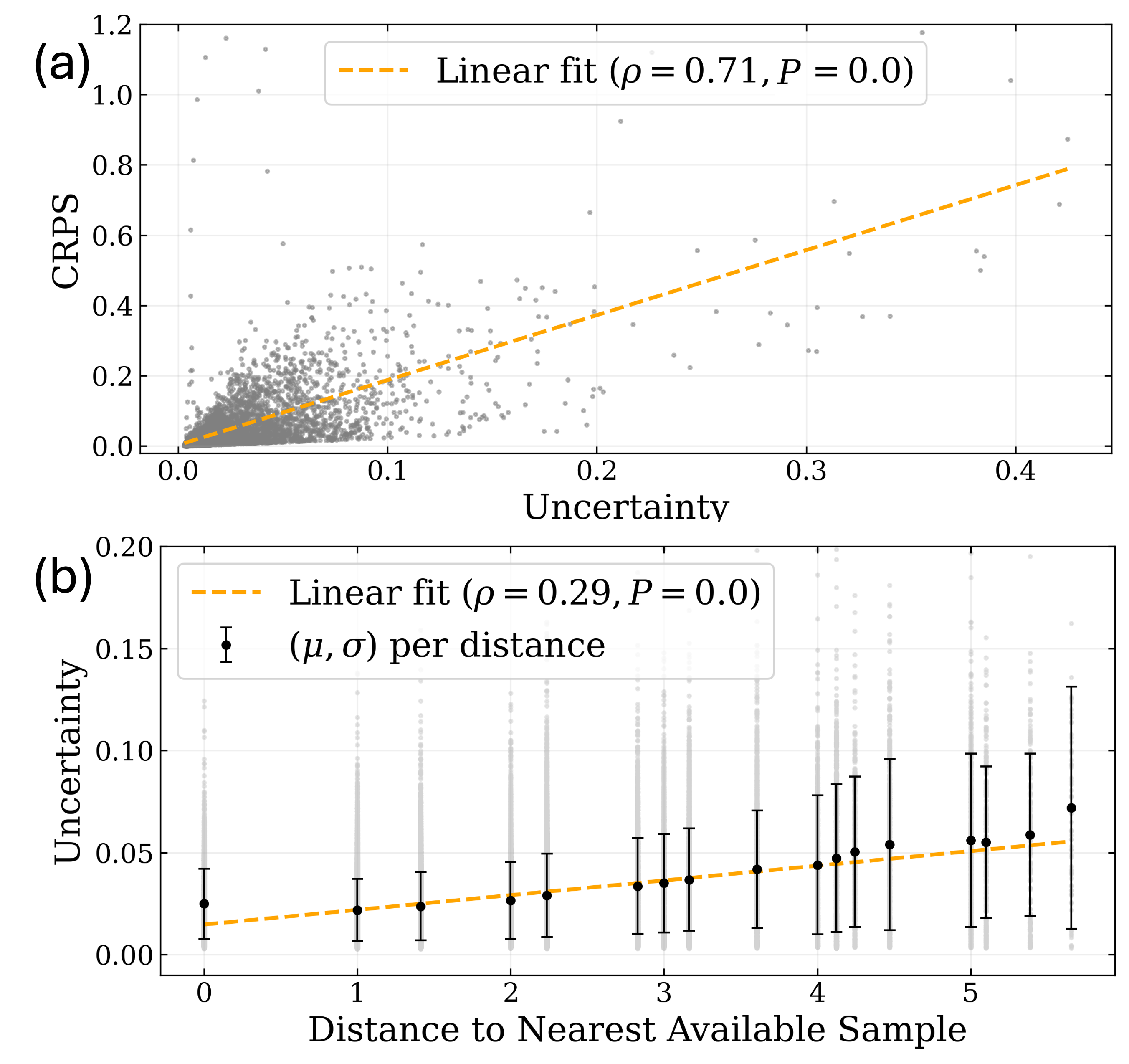}
\caption{\textbf{Uncertainty Calibration and Sample Coverage.} (a) Per-sample uncertainty correlates with error in CRPS. (b) Uncertainty heightens with distance to the nearest observed location.
}
\label{fig:uq_plots}
\end{center}
\vspace*{-0.25in}
\end{figure}

We assess whether SOLID’s predictive dispersion is informative and calibrated by turning DDIM ensembles into pixelwise uncertainty maps and then relating these maps to error, coverage, and forecast horizon. Concretely, for each sparse-conditioned query we draw $K{=}100$ conditional samples and compute the per-pixel standard deviation (Error) as our uncertainty estimate (Eq.~\ref{eq:uq}), using the same strict mask conditioning as in training; this yields an uncertainty map without ever supervising unobserved targets. Quantitatively, SOLID’s uncertainties are well calibrated: at the per-sample level, larger uncertainty coincides with worse CRPS (strong positive correlation), and uncertainty spatially increases with distance to the nearest observed location, mirroring the intuition that epistemic uncertainty grows in the ``void'' away from sensors (Fig.~\ref{fig:uq_plots}). This calibration is also reflected qualitatively: regions of high uncertainty co-locate with large error on Navier–Stokes (Fig.~\ref{fig:uq_qualitative}a) and with gaps in coverage on AirDelhi (Fig.~\ref{fig:air_delhi_perf}b), and during multi-step rollouts the uncertainty propagates and inflates as the prediction horizon extends (Fig.~\ref{fig:uq_qualitative}b). Together with the global observation that SOLID’s uncertainty maps are highly correlated with prediction error ($\rho{>}0.7$, Fig.~\ref{fig:uq_plots}), these results indicate that the model’s mask-conditioned diffusion not only reconstructs plausible fields but also provides actionable, calibrated confidence that tracks both task difficulty and data support, furnishing useful diagnostics for downstream assimilation and sensor-placement decisions.

\subsection{Task 4: Ablation Study}
\label{sec:ablations}
We ablate the overlap–reweighting coefficient $\lambda$ in the masked denoising loss. As shown in Fig.~\ref{fig:ablation_lambda}, the CRPS curve exhibits a clear optimum in the small–$\lambda$ regime and degrades when $\lambda$ is set either too small (under–uses conditioning) or too large (biases toward persistence), matching the intent of our overlap reweighting at $M_i \cap M_o$ described in Section~\ref{sec:method_sparse}. Complementing this, Table~\ref{tab:strategy-table} compares training strategies under varying input/target availability: training SOLID end–to–end \emph{without} any pre–interpolation attains the lowest errors, both at input locations ($M_i$) and away from the given locations, relative to Radial Basis Function (RBF) and nearest neighbor (NN) pre–interpolations. Together, these results justify using a modest overlap upweighting and strict mask–conditioned training without surrogate preprocessing, improving stability where inputs and targets overlap while preserving generalization beyond the conditioned regions.

\subsection{Task 5: Sparsity Patterns on Performance}
\label{sec:sparsity_study}

In practical lab setups and real sensor networks, one often faces highly sparse measurements due to limited instruments and sensor dropout (e.g., temporary failures or communication loss). To emulate these conditions, we systematically stress-test SOLID under varying sparsity regimes. Specifically, we vary three key factors: the overall sensor density (from extreme cases of 4\% coverage up to 40\%), the spatial structure of missing data (randomly scattered points vs. large contiguous missing blocks), and the sensor placement regime (fixed sensor locations across all instances vs. a new random mask per instance). This experimental design covers scenarios ranging from uniformly sparse sampling (mimicking random sensor outages) to worst-case structured gaps (entire regions devoid of sensors). The motivation is to assess how SOLID copes with both the level and pattern of sparsity—insights crucial for lab experiments where only a few sensors are available or when instruments drop out unexpectedly.

\begin{figure}[t]
\vspace*{-0.05in}
\begin{center}
\includegraphics[width=1.0\columnwidth]{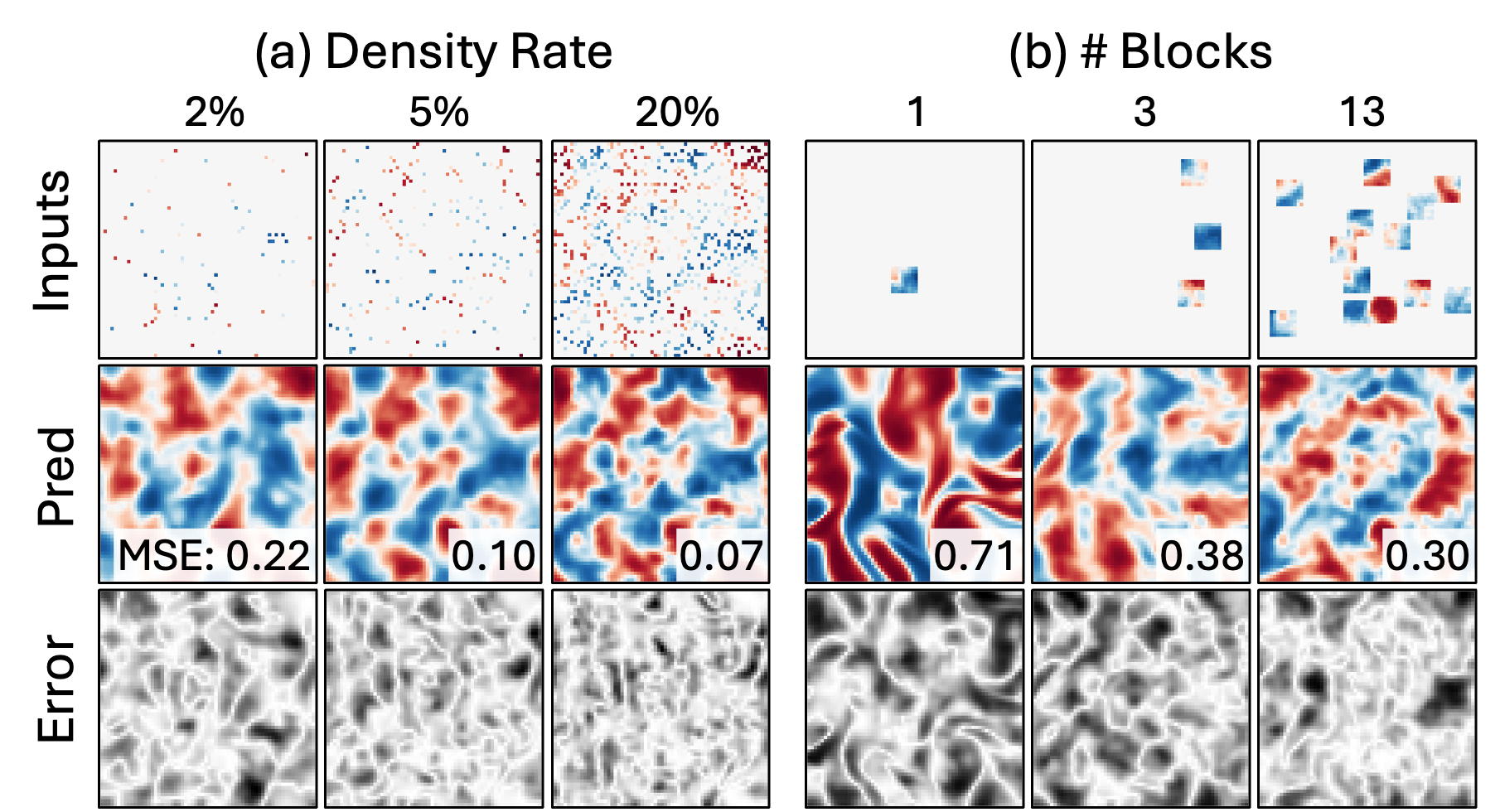}
\caption{\textbf{Sparsity Patterns on Performance.} (a) Increasing sensor density (2\%$\rightarrow$20\%) sharpens SOLID’s forecasting performance while error maps fade. (b) For a similar overall coverage, having sensors aggregated into condensed regions increases error.}
\label{fig:sparsity_pattern_isolated}
\end{center}
\vspace*{-0.2in}
\end{figure}

SOLID exhibits robust performance across all these challenging sparsity conditions, with clear trends emerging (Table~\ref{table:sparse_pattern}, Fig.~\ref{fig:sparsity_pattern_isolated}). Higher sensor density greatly improves accuracy: for randomly scattered sensors, SOLID’s CRPS improves by an order of magnitude as coverage increases from 4\% to 40\% (from roughly 1.3 to 0.3 under a fixed sensor layout, and from 1.2 to 0.1 when sensor positions vary per sample). In practical terms, even a modest addition of sensors yields disproportionately large gains in fidelity. The spatial pattern of sensor gaps also matters: at equal coverage, randomly distributed sensors are much more informative than a single large void region. For example, at 10\% coverage, a uniform random mask achieves a CRPS around 0.5, whereas concentrating sensors into one region (leaving a broad void area) can quadruple the error. Similarly, a single 8×8 sensor block yields high prediction error (e.g. 0.7 MSE in Fig.~\ref{fig:sparsity_pattern_isolated}), while using many smaller blocks (e.g. 13 blocks covering 40\% of the area) reduces error to 0.3 – approaching the performance of random coverage. 

We also find that varying the sensor layout between instances ($S_{\text{instance}}$) generally boosts performance (Table~\ref{table:sparse_pattern}). Training with a new random mask each time exposes the model to diverse observations and consistently lowers error compared to a fixed layout ($S_{\text{global}}$, for instance, at 10\% density the per-instance random strategy achieves CRPS 0.48~vs.~0.67 with a fixed mask). An interesting exception occurs under extremely sparse, highly localized sensing (only two small blocks of sensors): here a globally fixed layout slightly outperforms a varying one (CRPS 1.37~vs.~2.06), likely because a fixed placement guarantees at least some region is always observed. 

These findings translate into concrete guidance for deploying sparse sensors in real systems. To maximize reconstruction quality, practitioners should distribute sensors to cover the domain broadly and avoid large contiguous gaps in coverage. Whenever possible, introducing mobility or variability in sensor positions (e.g. rotating sensor locations or using a moving sensor) can significantly enhance performance by filling in different parts of the space over time. In summary, investing in even a few additional sensors and diversifying their placement yields substantial improvements in accuracy and uncertainty estimation, ensuring that models like SOLID remain reliable even under severe sparsity.




\begin{table}[t]
\centering
\small
\setlength{\tabcolsep}{5pt}
\renewcommand{\arraystretch}{0.9}
\caption{Comparison of strategies under sparse conditioning on Navier--Stokes. We report the performance at input (conditioned) locations ($M_i$), target-only locations ($M_o \setminus M_i$), and on the full grid ($U^{t_{o}}$). SOLID (no pre-interpolation) attains the lowest error across all three views. (\emph{Interp}: Pre-Interpolation; \emph{NN}: Nearest Neighbor; \emph{RBF}: Radial Basis Function)}
\begin{tabular}{lccc}
\toprule
\multirow{2}{*}{\textbf{Strategy}} & \multicolumn{3}{c}{\textbf{CRPS}} \\
\cmidrule(lr){2-4}
 & \textbf{$M_i$} & \textbf{$M_o \setminus M_i$} & \textbf{$U^{t_{o}}$} \\
\midrule
Pre-Interp (NN)   & 0.169 & 0.229 & 0.222 \\
Pre-Interp (RBF)  & \underline{0.081} & \underline{0.139} & \underline{0.133} \\ \cmidrule(lr){1-1}
SOLID             & \textbf{0.025} & \textbf{0.066} & \textbf{0.069} \\
\bottomrule
\end{tabular}
\label{tab:strategy-table}
\end{table}

\begin{table}[t]
\centering
\small
\caption{CRPS for Navier--Stokes experiments under different sparsity patterns and fixing regimes.}
\label{table:sparse_pattern}
\begin{tabular}{llcc}
\toprule
\multirow{2}{*}{\textbf{Pattern}}& \multirow{2}{*}{\textbf{Sample Density}}  & \multicolumn{2}{c}{\textbf{Sensor Scenarios}} \\
\cmidrule(lr){3-4}
&& \texttt{$S_\text{Global}$} & \texttt{$S_\text{Instance}$} \\
\midrule
\multirow{3}{*}{Random}
  & 4\%   & 1.37 & 1.23 \\
  & 10\%   & \underline{0.67} & \underline{0.48} \\
  & 40\%  & \textbf{0.31} & \textbf{0.10} \\
\midrule
\multirow{3}{*}{Block (8$\times$8)}
  & 2 blocks   & 1.37  & 2.06 \\
  & 6 blocks   & 2.04 & 1.84  \\
  & 26 blocks  & 0.97 & 0.95 \\
\bottomrule
\end{tabular}
\vspace*{-0.15in}
\end{table}


\section{Conclusion}
\label{sec:discussions}

\paragraph{Limitations \& Future Work.}
While SOLID is parameter-efficient per denoising step, sampling remains expensive at inference as we run 50 denoising steps, and uncertainty estimation further requires multiple Monte Carlo samples. Thus, our empirically calibrated uncertainty from sparse observations comes with a clear trade-off: improved fidelity and uncertainty quantification at increased compute. Future work will focus on accelerating sampling (e.g., fewer-step samplers / distillation) and extending the framework to multimodal spatiotemporal data by integrating heterogeneous sensor types and data sources.

We present SOLID, a conditional probabilistic diffusion model that reconstructs and forecasts spatiotemporal fields directly from sparse sensor measurements. Across diverse settings, SOLID appears parameter- and data-efficient. Its dual-masking strategy and sparsity-aware objective reliably recover unobserved regions and yield calibrated ensemble predictions that remain physically plausible. The derived uncertainty is correlated with error, sampling locations, and prediction time horizon, providing deep insights on forecasting. These properties make SOLID a practical foundation for learning from real-world sensor networks.

\section*{Impact Statement}

This paper advances machine learning methods for reconstructing and forecasting spatiotemporal physical fields from sparse observations, while providing uncertainty estimates. Potential benefits include improved monitoring and decision support in settings such as environmental sensing and scientific modeling when dense ground truth is unavailable. Potential risks arise from misuse or overreliance: errors or miscalibrated uncertainty could lead to poor decisions, and uneven sensor coverage may introduce systematic biases. Diffusion-based sampling can also be computationally costly. We recommend validation, calibration checks, and clear communication of uncertainty and limitations for any real-world deployment.

\section*{Acknowledgements}
This work was supported by the U.S. Department of Energy (DOE), Office of Science (SC), Advanced Scientific Computing Research program under award DE-SC-0012704 and used resources of the National Energy Research Scientific Computing Center (NERSC), a Department of Energy User Facility using NERSC award ASCR-ERCAP0035391.



\bibliography{main}
\bibliographystyle{icml2026}

\newpage
\appendix
\onecolumn

\section{Dataset Background and Processing}
\label{app:dataset}


\subsection{Navier--Stokes dataset}

This dataset is based on an incompressible fluid flow governed by the vorticity transport equation
\begin{equation*}
    \frac{\partial \omega}{\partial t}
    = -\mathbf{u} \cdot \nabla \omega
      + \nu \Delta \omega
      + f,
\end{equation*}
where the vorticity is defined by
\begin{equation}
    \omega = \nabla \times \mathbf{u}, \qquad \nabla \cdot \mathbf{u} = 0.
\end{equation}
Here, \( \mathbf{u} \) denotes the velocity field and \( \omega \) the associated vorticity, both specified on a spatial domain with periodic boundary conditions. The parameter \( \nu \) is the kinematic viscosity, and \( f \) is an external forcing term used to sustain turbulence.

At each time \( t \), the input to the model is given by \( v_t = \omega_t \). The spatial domain is
\[
    \Omega = [-\pi, \pi]^2.
\]
The initial vorticity field is drawn from a Gaussian Random Field (GRF) with parameters
\[
    \alpha = 2.5, \qquad \tau = 3.0.
\]
The parameter \( \alpha \) controls the smoothness of the initial vorticity field, while \( \tau \) sets the characteristic correlation length of spatial structures.

The external forcing used in the simulation is
\begin{equation}
    f(x_1, x_2) = -4 \cos(4x_2),
\end{equation}
where \( x_2 \) is the vertical spatial coordinate. This term induces a vertically periodic forcing pattern that promotes rotational motion. The periodicity of the cosine function creates a repeating vortex structure that maintains turbulence and counteracts the decay of kinetic energy. The negative sign ensures a consistent direction of vorticity injection, reinforcing the rotational dynamics of the flow. As a consequence, the system exhibits a persistent and well-organized turbulent pattern.

We simulate the flow at a Reynolds number \( Re = 100 \), corresponding to a moderately turbulent regime in which inertial effects dominate over viscous dissipation, enabling complex vortex interactions while retaining numerical stability. This setup is particularly suitable for spatiotemporal learning tasks, since it generates a rich yet structured temporal evolution of the vorticity field, providing an effective testbed for models that aim to learn continuous representations of dynamical physical systems.

In total, we generate \(100{,}000\) trajectories, each with \(T = 50\) time steps. For training and validation, we use every other time step,
\[
    \{0, 2, 4, \dots, 48\},
\]
while the remaining time steps,
\[
    \{1, 3, 5, \dots, 49\},
\]
are reserved to evaluate the model’s ability to infer continuous-time dynamics from discrete observations. We plan to release our simulated data to readers for future evaluation purposes.





\subsection{AirDelhi dataset}
The AirDelhi~\citep{airdelhi} dataset provides a detailed collection of fine-grained spatiotemporal particulate matter (PM) measurements from Delhi, India, obtained by mounting low-cost PM${}{2.5}$ sensors on 13 public buses operating across the Delhi-NCR region (covering roughly $559~\text{km}^2$) to overcome the constraints of static sensor networks. In total, the campaign comprises about $12.5$ million ground-level samples collected over 91 days (November~1, 2020 to January~31, 2021), with each bus typically operating for $16$–$20$ hours per day and reporting measurements at a resolution of 20 samples per minute. Each record contains 10 variables, including three pollutant measurements (PM${}1$, PM${}{2.5}$, PM${}{10}$) and seven contextual attributes (latitude, longitude, timestamp, device identifier, pressure, temperature, and relative humidity), with essentially no missing values for the PM fields; over the campaign period, mean PM concentrations are high (approximately $120~\mu\text{g/m}^3$ for PM${}1$, $208~\mu\text{g/m}^3$ for PM${}{2.5}$, and $226~\mu\text{g/m}^3$ for PM${}_{10}$) with large standard deviations, reflecting substantial spatiotemporal variability in Delhi’s air quality. While the raw data provide dense measurements along bus routes, the movement of sensors leads to sparse and temporally inconsistent observations at any fixed location, requiring models that can effectively process data from sparsely and dynamically moving sensors.

For our benchmark, we follow the preprocessing protocol in~\citet{airdelhi} and focus on a filtered subset of this campaign. Specifically, we consider data spanning November~12, 2020, to January~30, 2021, excluding the initial days with limited samples and fewer bus-mounted instruments, as well as nighttime intervals from 10{:}00~PM to 5{:}30~AM when buses are parked at depots. The study area is partitioned into $1~\text{km}^2$ spatial grids, which are further divided into spatiotemporal cells with 30-minute intervals, and the mean of all measurements within each cell is used as the representative PM value. We adopt the \emph{AB} and \emph{CP} subsets as training and test sets, respectively, following the original split. This benchmark is characterized by differing numbers of samples for the input ($N_i$) and output ($N_o$) in each instance, which, together with the sparsity induced by the moving sensors, places strong demands on model flexibility.

\subsection{Construction of Sparse Observation Scenarios for the Navier--Stokes Dataset}
\label{sec:sparsity_navier_stokes}

In all experiments, the underlying data consist of fully resolved two-dimensional
Navier--Stokes snapshots on a \(64 \times 64\) spatial grid. The simulated sparsity
never alters the underlying simulation; instead, it is implemented by applying
binary masks to these dense fields.

For every training example, we construct two binary masks on the grid:
\begin{enumerate}
    \item a \emph{conditioning mask}, indicating the grid points where the true
    field values are revealed to the model and used as input;
    \item a \emph{target mask}, indicating the grid points where the loss is
    computed and predictions are evaluated.
\end{enumerate}
The union of these two masks defines the total ``sensor coverage'' or
``density'' for a given scenario. The reported densities (4\%, 10\%, 40\%)
correspond to the fraction of grid points that are used by either the
conditioning or the target mask. In most configurations we consider, the total
number of selected locations is split approximately in half between the two
masks: about half of the selected locations belong to the conditioning mask and
the other half to the target mask. In the general framework, the two sets may
overlap; locations in the intersection are simultaneously used for conditioning
and supervision (and can be reweighted accordingly in the loss), while the
non-overlapping parts are used exclusively for input or target.

During training, for a given example, the dense input field is multiplied by the
conditioning mask so that only conditioning locations are visible to the model;
the loss is then applied only on the target-mask locations (with a possible
additional weighting on overlapping locations). All other grid points are
treated as missing: they are neither provided as input nor directly supervised.

Each sparsity scenario is defined along two axes:
\begin{enumerate}
    \item the \emph{spatial pattern} of observed locations:
    \begin{itemize}
        \item \textbf{Random}: individual pixels are randomly scattered across
        the grid;
        \item \textbf{Block}: \(8 \times 8\) square blocks are
        placed on the grid and used in their entirety (subject to a
        non-overlap constraint between blocks).
    \end{itemize}
    \item the \emph{variation across examples}:
    \begin{itemize}
        \item \textbf{Global} (\(S_{\text{global}}\)): one pair of masks is
        drawn once and used for every example in the dataset;
        \item \textbf{Instance-specific} (\(S_{\text{instance}}\)): each example
        receives its own pair of masks, which remains fixed for that example
        across training.
    \end{itemize}
\end{enumerate}
All scenarios below are obtained by combining these two axes with a chosen
coverage level. Unless otherwise noted, we choose the conditioning and target
subsets to be non-overlapping for clarity in the Navier--Stokes stress tests;
however, the same machinery can be configured to introduce a controlled amount
of overlap between input and target locations.

\subsubsection{Random Masks (Pixel-wise Sparsity)}

In the random scenarios, sparsity is defined at the level of individual pixels.
For a chosen density \(p \in \{4\%, 10\%, 40\%\}\), we first compute the number
of pixels this corresponds to on the \(64 \times 64\) grid (with 4096 pixels in
total). For example, a density of 4\% corresponds to approximately
\(0.04 \times 4096 \approx 163\) pixels, 10\% to approximately 409 pixels, and
40\% to approximately 1638 pixels.

For a given example, we select this number of pixels uniformly at random,
without replacement, and then partition the selected set into two subsets
(up to an off-by-one difference when the count is odd):
\begin{itemize}
    \item one subset is assigned to the conditioning mask (visible input points);
    \item the other subset is assigned to the target mask (supervised output
    points).
\end{itemize}
In the Navier--Stokes sparsity experiments, this partition is chosen so that
conditioning and target pixels are disjoint, which implies that in a 4\% random
scenario, roughly 2\% of the grid is used as conditioning and 2\% as targets;
in a 10\% random scenario, roughly 5\% conditioning and 5\% target; and in a
40\% random scenario, roughly 20\% conditioning and 20\% target. More generally,
the same construction can be adapted to allow part of the selected pixels to be
shared between conditioning and target (by assigning some locations to both
masks), in which case the overlap portion participates simultaneously in
conditioning and supervision and can be reweighted in the loss.

The distinction between global and instance-specific variants is how these
randomly chosen pixels are reused across the dataset:
\begin{itemize}
    \item \textbf{Random 4\%, global}:
    a single pixel set covering approximately 4\% of the grid is drawn once,
    partitioned into conditioning and target subsets, and the resulting pair of
    masks is applied to every snapshot in the dataset. Every example is observed
    at exactly the same sparse set of locations.
    \item \textbf{Random 4\%, instance-specific}:
    each example draws its own pixel set covering approximately 4\% of the
    grid, partitions it into conditioning and target subsets, and then reuses
    that pair only for that example. Different examples have different 4\%
    layouts, but each example's layout is stable across epochs.
    \item \textbf{Random 10\%, global}:
    a single 10\%-dense pixel set is drawn and partitioned between conditioning
    and target. All examples share the same 10\% layout.
    \item \textbf{Random 10\%, instance-specific}:
    each example has its own random 10\% layout, again partitioned roughly into
    equal conditioning and target subsets. The layout varies across examples but
    is fixed for each example over time.
    \item \textbf{Random 40\%, global}:
    one dense 40\% layout is drawn, partitioned into two subsets assigned to
    conditioning and target, and reused for every example.
    \item \textbf{Random 40\%, instance-specific}:
    each example has its own dense 40\% layout, partitioned into conditioning
    and target subsets and kept fixed for that example.
\end{itemize}

In summary, all random scenarios select a prescribed fraction of pixels, split
them into conditioning and target masks (with zero or controlled overlap), and
differ only in the overall density (4\%, 10\%, 40\%) and in whether the resulting
pattern is shared across all data (global) or defined separately for each example
(instance-specific).

\subsubsection{Block Masks (Structured \(8 \times 8\) Sparsity)}

In the block scenarios, sparsity is defined in terms of contiguous
\(8 \times 8\) patches (blocks) rather than individual pixels. Each block
contains 64 pixels. Blocks are placed at random locations on the \(64 \times 64\)
grid subject to a non-overlap constraint between blocks themselves.

The scenarios use a total of 2, 6, or 26 blocks. On the \(64 \times 64\) grid
(with 4096 pixels), these correspond to:
\begin{itemize}
    \item 2 blocks: \(2 \times 64 = 128\) pixels, about 3.1\% of the grid;
    \item 6 blocks: \(6 \times 64 = 384\) pixels, about 9.4\% of the grid;
    \item 26 blocks: \(26 \times 64 = 1664\) pixels, about 40.6\% of the grid.
\end{itemize}
For a given scenario, the total set of blocks is then split evenly between
conditioning and target masks:
\begin{itemize}
    \item with 2 blocks, 1 block is assigned to conditioning and 1 block to
    targets;
    \item with 6 blocks, 3 blocks are assigned to conditioning and 3 to targets;
    \item with 26 blocks, 13 blocks are assigned to conditioning and 13 to
    targets.
\end{itemize}
Within an example, the input field is visible only inside the conditioning
blocks, the loss is computed only inside the target blocks, and the remainder of
the domain is unobserved. As with the random case, the framework could be
extended to allow some blocks to be shared between conditioning and target
masks, but in the Navier--Stokes experiments we use disjoint assignments of
blocks to conditioning and target for simplicity.

Global and instance-specific variants again differ in how these blocks are
reused across examples:
\begin{itemize}
    \item \textbf{Block, 2 blocks, global}:
    two non-overlapping \(8 \times 8\) blocks are placed once on the grid;
    one block is designated as the conditioning region and the other as the
    target region. Every example uses exactly this same pair of large blocks:
    one fixed input island and one fixed supervised island, with the rest of
    the field unobserved.
    \item \textbf{Block, 2 blocks, instance-specific}:
    for each example, two non-overlapping \(8 \times 8\) blocks are placed at
    random locations, and then one is used as the conditioning block and the
    other as the target block for that example. Different examples see their
    own pair of large observed patches, but each example's pair remains fixed
    during training.
    \item \textbf{Block, 6 blocks, global}:
    six non-overlapping \(8 \times 8\) blocks are placed once. Three are
    assigned to conditioning and three to target, and this six-block pattern
    is shared by every example.
    \item \textbf{Block, 6 blocks, instance-specific}:
    each example places its own six non-overlapping \(8 \times 8\) blocks,
    then splits them into three conditioning and three target blocks. Layouts
    differ between examples but are consistent for a given example.
    \item \textbf{Block, 26 blocks, global}:
    twenty-six non-overlapping \(8 \times 8\) blocks are placed once. Thirteen
    are assigned to conditioning and thirteen to target, and this dense
    patchwork pattern is used for every example.
    \item \textbf{Block, 26 blocks, instance-specific}:
    each example uses its own set of 26 non-overlapping \(8 \times 8\) blocks,
    split into 13 conditioning and 13 target blocks. Each example thus has a
    dense but unique block layout, constant over time for that example.
\end{itemize}

In all block scenarios, the effective coverage (around 3\%, 9\%, or 40\%) is
approximately aligned with the 4\%, 10\%, and 40\% random scenarios, but the
information is concentrated in contiguous patches rather than scattered
individual pixels. Global variants correspond to a single fixed block layout for
the entire dataset, whereas instance-specific variants correspond to a different
block layout for each example. When overlap between conditioning and target is
enabled at the block level, any shared blocks would provide both input and
supervision and can be given special weight in the loss.

\subsubsection{Unified View}

Overall, every sparsity scenario is constructed by:
\begin{enumerate}
    \item choosing a total coverage level (specified directly as a percentage of
    pixels or indirectly via the number of \(8 \times 8\) blocks);
    \item splitting that coverage into conditioning and target regions of
    roughly equal size, optionally allowing some locations or blocks to be
    shared between the two masks; and
    \item deciding whether the resulting layout is shared globally across the
    dataset (\(S_{\text{global}}\)) or fixed separately for each individual
    example (\(S_{\text{instance}}\)).
\end{enumerate}
This framework yields a family of sparse observation regimes that are
systematically comparable in terms of density, spatial structure, and the degree
of overlap between input and target samples.

\begin{table}[t]
\centering
\caption{\textbf{Compute and model size.} FLOPs are reported in GFLOPs.}
\label{tab:flops_params_all}
\setlength{\tabcolsep}{7pt}
\renewcommand{\arraystretch}{1.08}
\begin{tabular}{l l S[table-format=5.3] S[table-format=8.0]}
\toprule
\textbf{Model} & \textbf{Basis} & {\textbf{FLOPs (GFLOPs)}} & {\textbf{\# Params}} \\
\midrule
SCENT     & per forward pass    & 20.351     & 28988739 \\
OmniField & per forward pass    &  9.966     & 10272327 \\
FFNO      & per forward pass    &  0.771     &   535746 \\
OFormer   & per forward pass    &  6.682     &  2662849 \\
UNet     & per forward pass    &  0.397     &   466593 \\
StyleGAN3 & per forward pass    & 10473.000  & 76320261 \\
INR-GAN   & per forward pass    & 2147.000   & 55300000 \\
\midrule
DDO          & per denoising step & 23.723    & 11103780 \\
InfinityDiff & per denoising step &  6.790    & 40909722 \\
SOLID        & per denoising step &  5.919    &  8950913 \\
\bottomrule
\end{tabular}
\end{table}



\begin{table*}[t]
\centering
\caption{\textbf{SOLID hyperparameters.} Navier--Stokes values are from the provided config. AirDelhi values are from the provided training cell.}
\label{tab:solid_hparams_appendix}
\setlength{\tabcolsep}{7pt}
\renewcommand{\arraystretch}{1.08}
\begin{tabular}{lcc}
\toprule
\textbf{Hyperparameter} & \textbf{Navier--Stokes} & \textbf{AirDelhi} \\
\midrule
\multicolumn{3}{l}{\textit{Model}} \\
\midrule

Backbone & UNet & UNet \\
UNet base dim & 64 & 64 \\
Image size & 64 & $21 \times 20$ \\
Dim multipliers & (1, 2, 2, 2) & (1, 2, 2,2) \\
Attention resolutions & (16) & (16) \\
Dropout & 0.1 & 0.1 \\
ResNet blocks / stage & 2 & 2 \\

\midrule
\multicolumn{3}{l}{\textit{Training}} \\
\midrule
Dataset & NavierStokes & AirDelhi \\
Batch size & 64 & 32 \\
Optimizer & AdamW & AdamW \\
Learning rate & $2 \times 10^{-4}$ & max $4 \times 10^{-4}$, min $8 \times 10^{-6}$ \\
LR schedule & Cosine Annealing & Cosine Annealing \\
Weight decay & $1 \times 10^{-4}$ & $1 \times 10^{-4}$ \\
Total training steps & 600{,}000 & 150{,}000 \\
Gradient clipping & enabled & enabled \\

\midrule
\multicolumn{3}{l}{\textit{Sampling}} \\
\midrule
Sampler & DDIM & DDIM \\

DDIM sampling steps & 50 & 50 \\

\bottomrule
\end{tabular}
\end{table*}

\begin{table}[t]
  \captionsetup{font=small}
  \caption{\textbf{Full Results on Navier--Stokes across sparsity} (CRPS, corresponds to Fig.~\ref{fig:efficiency}).
  Best per column in bold.}
  \label{tab:navierstokes-crps}
  \small
  \centering
  \begin{tabular}{lrrrrrr}
    \toprule
    & \multicolumn{3}{c}{\textbf{Full}} & \multicolumn{3}{c}{\textbf{10\% Subset}} \\
    \cmidrule(lr){2-4}\cmidrule(lr){5-7}
    \textbf{Baseline} & \textbf{4\%} & \textbf{10\%} & \textbf{40\%} & \textbf{4\%} & \textbf{10\%} & \textbf{40\%} \\
    \midrule
    SCENT         & 1.6638 & 0.8490 & 0.5568 & 1.9158 & 1.3971 & 1.2591 \\
    OmniField     & 1.6982 & 1.1019 & 0.4765 & 2.1117 & 1.6541 & 1.0651 \\
    FFNO          & 1.6863 & 1.0993 & 0.4545 & 1.8689 & 1.3265 & 0.7042 \\
    OFormer       & 1.6873 & 1.0358 & 0.5645 & 1.9677 & 1.2387 & 0.9052 \\
    UNet           & 1.6256 & 0.9981 & 0.4673 & 1.8204 & 1.2387 & 0.6722 \\
    SOLID         & 1.2141 & \textbf{0.3840} & \textbf{0.0968} & \textbf{1.2251} & \textbf{0.4826} & \textbf{0.1005} \\
    StyleGAN3     & 2.6558 & 2.3733 & 2.1760 & 2.6379 & 2.6364 & 2.6354 \\
    INR-GAN       & 2.5673 & 2.4317 & 2.0803 & 2.6293 & 2.5688 & 2.5478 \\
    DDO           & \textbf{1.1016} & 0.9924 & 0.8108 & 1.2891 & 1.2879 & 1.2803 \\
    $\infty$-Diff & 1.4419 & 0.7276 & 0.2106 & 1.7660 & 1.2084 & 0.4610 \\
    \bottomrule
  \end{tabular}
\end{table}

\begin{table}[t]
  \centering
  \caption{\textbf{Full Results on AirDelhi} (CRPS, corresponds to Fig.~\ref{fig:air_delhi_perf}).}
  \label{tab:airdelhi}
  \begin{tabular}{l S}
    \toprule
    \multicolumn{1}{c}{\textbf{Baseline}} &
    \multicolumn{1}{c}{\textbf{AD-B}} \\
    \midrule
    
    SOLID                 & \textbf{25.713} \\
    SCENT          & 28.467 \\
    OmniField    & 29.244 \\
    FFNO        & 30.156       \\
    UNet         & 32.897   \\
    OFormer      & 57.837  \\
    DDO        & 77.598  \\
    INR-GAN     & 39.383     \\
    StyleGAN3  &  33.135      \\
    $\infty$-Diff        & 36.120   \\

    \bottomrule
  \end{tabular}
\end{table}

\end{document}